\documentclass[letterpaper]{article} % DO NOT CHANGE THIS
\usepackage{aaai25}
\usepackage{times}  % DO NOT CHANGE THIS
\usepackage{helvet}  % DO NOT CHANGE THIS
\usepackage{courier}  % DO NOT CHANGE THIS
\usepackage[hyphens]{url}  % DO NOT CHANGE THIS
\usepackage{graphicx} % DO NOT CHANGE THIS
\usepackage{natbib}  % DO NOT CHANGE THIS AND DO NOT ADD ANY OPTIONS TO IT
\usepackage{caption} % DO NOT CHANGE THIS AND DO NOT ADD ANY OPTIONS TO IT
\usepackage{algorithm}
\usepackage{algorithmic}
\usepackage{caption}
\usepackage[subrefformat=simple]{subcaption} %図の説明を書くために追加した
\usepackage{here} %好きな位置に図を置くため
\usepackage{bm}
\usepackage{amsmath}
\usepackage{multirow}
\usepackage{comment}  %自分で追加した
\usepackage{textcomp}
\usepackage{mathcomp}
\usepackage{enumerate}
\usepackage{amsfonts}
\usepackage{newfloat}
\usepackage{listings}
\DeclareCaptionStyle{ruled}{labelfont=normalfont,labelsep=colon,strut=off} % DO NOT CHANGE THIS
\floatstyle{ruled}
\newfloat{listing}{tb}{lst}{}
\floatname{listing}{Listing}
\title{Multi-instance Learning as Downstream Task of Self-Supervised Learning-based Pre-trained Model}
\author{
    Koki Matsuishi\textsuperscript{\rm 1}, Tsuyoshi Okita\textsuperscript{\rm 1}
}
\affiliations{
    \textsuperscript{\rm 1}
    Kyushu Institute of Technology
680-4 Kawazu, Iizuka-shi, Fukuoka, 820-8502, JAPAN\\
    matsuishi.kouki587@mail.kyutech.jp, \\
    tsuyoshi@ai.kyutech.ac.jp
}

\usepackage{bibentry}
\begin{document}

\maketitle

\begin{abstract}
In deep multi-instance learning, the number of applicable instances depends on the data set. In histopathology images, deep learning multi-instance learners usually assume there are hundreds to thousands instances in a bag. However, when the number of instances in a bag increases to 256 in brain hematoma CT, learning becomes extremely difficult. In this paper, we address this drawback. To overcome this problem, we propose using a pre-trained model with self-supervised learning for the multi-instance learner as a downstream task. With this method, even when the original target task suffers from the spurious correlation problem, we show improvements of 5\% to 13\% in accuracy and 40\% to 55\% in the F1 measure for the hypodensity marker classification of brain hematoma CT.
% AAAI creates proceedings, working notes, and technical reports directly from electronic source furnished by the authors. To ensure that all papers in the publication have a uniform appearance, authors must adhere to the following instructions.
\end{abstract}


\section{Introduction}
Multiple Instance Learning (MIL)\cite{DIETTERICH199731,NIPS1997-82965d4e} is a type of weakly supervised learning. In MIL, a collection of instances is defined as a "bag," and learning is performed by referencing the labels of these bags. Recently, Attention-based Deep Multiple Instance Learning(Deep MIL)\cite{ITW:2018} has been proposed, making deep learning-based MIL approaches mainstream. In traditional MIL settings, the number of instances within a bag is often limited to a few dozen due to optimization constraints. However, with deep learning-based MIL approaches, it has become possible to handle hundreds to thousands of instances. In fact, experiments in Deep MIL involving histopathology images included over 600 instances per bag. Histopathology images are a major application area for MIL, and due to the high resolution of these images, they often contain several hundred to several thousand instances (patch images)\cite{shao2021transmil,lin2023interventional,li2021dual}. \\ \indent However, in this study, we focus on a marker classification task for brain hematoma CT images, which differs from the classification of histopathology images. Brain hematomas can be categorized into rapidly growing and non-rapidly growing types, and markers have been devised to distinguish between them\cite{pmid27323314}. One such marker is known as hypodensity, and the classification of these markers is recognized as a challenging task due to spurious correlation problem. We applied Deep MIL by dividing brain hematoma CT image slices into 256 patch images, resulting in 256 instances. The intention behind dividing the images into smaller patch images was to enable discrimination based on image texture. However, this approach resulted in poor performance, and the attention mechanism failed to capture the hematomas effectively. We found that MIL with 256 instances is challenging for our target task of brain hematoma marker classification. \\
\indent To overcome this problem, we propose using a pre-trained model with self-supervised learning for the multi-instance learner as a downstream task. Specifically, by using patch images created from divided CT slice images as input, we employed self-supervised learning based on contrastive learning and reconstruction tasks. The trained encoder was then used as an instance feature extractor in downstream MIL. This provided effective representation information for the patch images, making it easier to perform MIL even with 256 instances. In summary, our contributions are as follows:
\begin{itemize}
\item We propose using a pre-trained model with self-supervised learning for the deep multi-instance learner as a downstream task. Self-supervised learning of patch images provides information on valid representations to the instance feature extractor.
\item In the classification of marker in CT brain hematoma images, which is one of the difficult tasks due to the spurious correlation problem, MIL can be performed despite the large number of instances.
\item Experimental results show superior performance compared to methods that apply Deep MIL without self-supervised learning.
\end{itemize}

\section{Related Work} 
\subsection{Multiple Instance Learning with Self-supervised Learning}
\indent For multiple instance learning in medical imaging, methods applying self-supervised learning to instance feature extractors have been proposed. Many studies applying multiple instance learning to medical images use models pre-trained on ImageNet~\cite{5206848} for instance feature extractors, but due to the differences between medical and natural images, these may not be optimal for obtaining useful feature representations of medical images. Since it is often difficult to obtain instance-level label information in medical images for supervised learning, self-supervised learning can be applied for pre-training the instance feature extractors. In studies focusing on histopathology images, methods applying SimCLR~\cite{silva2020exploringsimclr}, models combining VAE (Variational Autoencoder)~\cite{Kingma2014} and GAN (Generative Adversarial Networks)~\cite{goodfellow2014generative}, methods applying MoCo v2~\cite{Chen2020ImprovedBW}, and methods applying DINO~\cite{caron2021emerging} with ViT~\cite{dosovitskiy2020vit} as the backbone have been proposed. On the other hand, studies focusing on CT images include methods that apply tasks predicting the absolute and relative positions of 12 patch images divided from slice images for severity assessment of COVID-19 in chest CT images~\cite{LI2021101978}, and methods predicting which of three different scale transformations was applied to the slice images for detecting HCC (hepatocellular carcinoma) in liver CT images~\cite{62d0cd7420374a2e99113e71593665f5}. Our brain hematoma CT image dataset and the marker classification task for it differ in setting and context from these histopathology and CT image examples. The spurious correlation problem is one of them, and these methods are not necessarily applicable to our task. Also, as far as we know, there are no studies applying self-supervised learning to instance feature extractors for classification of brain hematoma CT images.

\section{Methology} \label{method}
\begin{figure*}[t]
\begin{center}
\includegraphics[width=15cm]{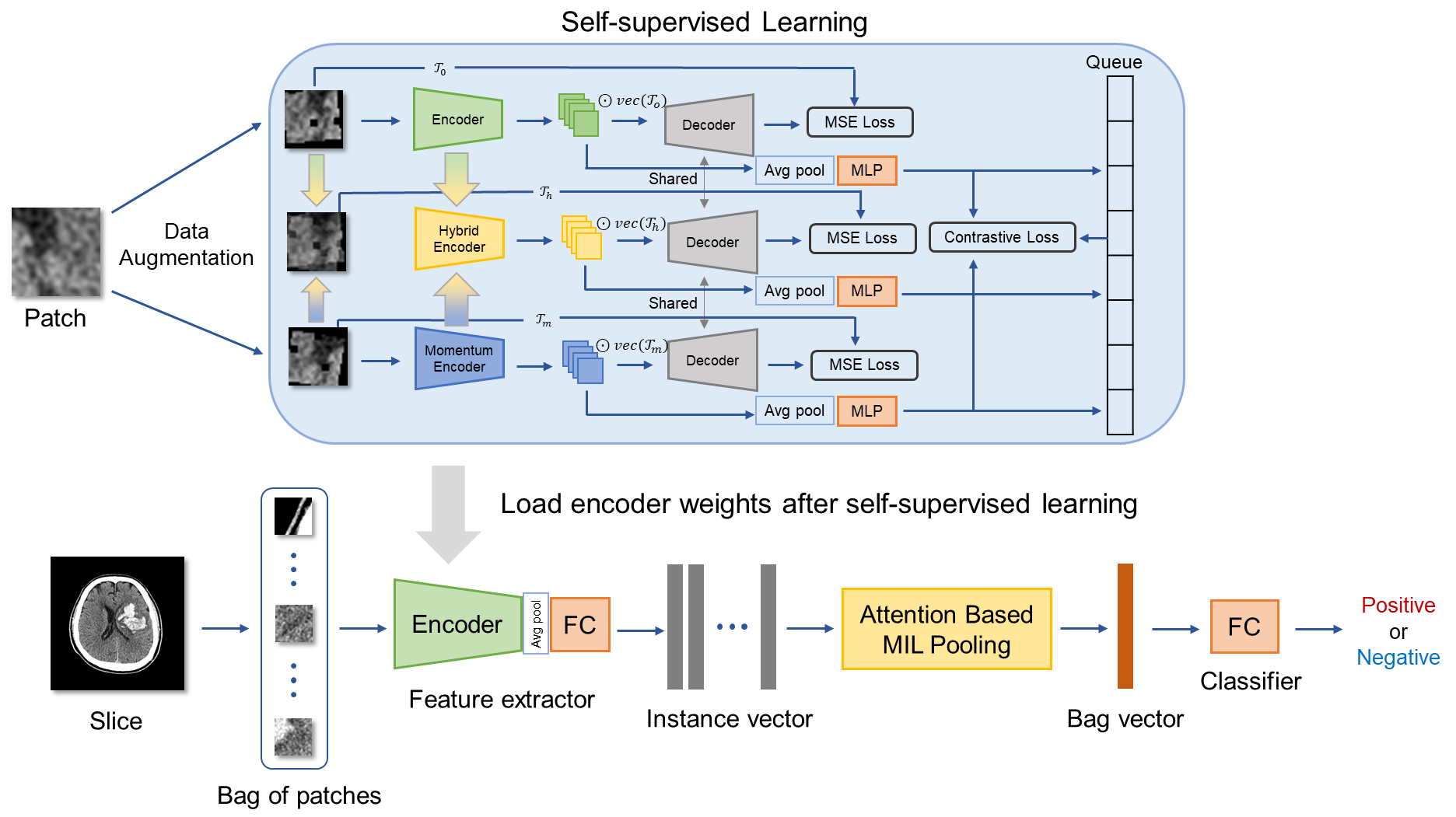}
\caption{Overview of our method. In the first stage, self-supervised learning is performed using patch images created by dividing the slices into a grid as input; in the second stage, deep multi-instance learning is performed using CT slices as bags and patch images as instances. The feature extractor for instances uses the encoder weights that have been pre-trained by self-supervised learning.}
\label{fig:method}
\end{center}
\end{figure*}

In this paper, we propose a method for classifying brain hematoma marker in brain CT slice images using Deep MIL, with the instance feature extractor pretrained using self-supervised learning on patch images. An overview of our method is shown in Figure \ref{fig:method}.
\begin{figure*}[t]
\begin{center}
\includegraphics[width=15cm]{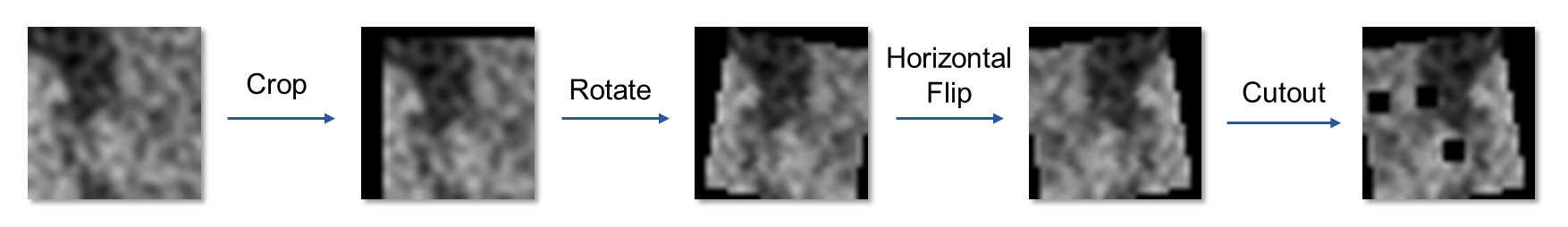}
\caption{Examples of data augmentation on patch images}
\label{fig:DA}
\end{center}
\end{figure*}
\subsection{Multiple Instance Learning Formulation}
In MIL, label information for individual instances is unavailable during training, but the label information for a bag (a collection of instances) is accessible. MIL is often set up as a binary classification problem. In this case, a bag is considered positive if it contains at least one positive instance, and negative if it contains only negative instances. The bag's label does not depend on the order of the instances, and the number of instances in a bag does not have to be the same across all bags. In the brain CT images used in this study, the slices are bags, and each patch image created by dividing the slices into a grid corresponds to an instance.\\
Let there be an instance $\bm{x}\in\mathbb{R}^{D}$ with a label $y\in\lbrace0,1\rbrace$, which cannot be referenced during training. For a bag $X=\lbrace \bm{x}_{1},\bm{x}_{2},...,\bm{x}_{K}\rbrace$ composed of $K$ instances, where each instance has a label $\lbrace y_{1},y_{2},...,y_{K}\rbrace$, the label of the bag $Y$ is given by  (\ref{label_formal}).
\begin{align}
\label{label_formal}
Y
 =
  \begin{cases}
    0, & \mathrm{if}\,\sum_{k=1}^{K}y_{k}=0, \\
    1, & \mathrm{otherwise.}
  \end{cases}
\end{align}
% Additionally, since the label $Y$ is independent of the order of instances in the bag $X=\lbrace \bm{x}{1},\bm{x}{2},...,\bm{x}_{K}\rbrace$, it can also be expressed as in  (\ref{label_another}).
% \begin{align}
% \label{label_another}
% Y=\underset{k}{\mathrm{max}}\lbrace y_{k}\rbrace.
% \end{align}
\indent A common approach to classify a bag involves three stages: 1) transforming individual instances using a function $f(\cdot)$, 2) pooling the transformed instances using a identity(order-invariant) function $\sigma(\cdot)$, and 3) transforming the pooled instances using a function $g(\cdot)$. The predicted label $\hat{Y}\in\lbrace0,1\rbrace$ for the bag is given by  (\ref{bag_classifier}).
\begin{align}
\label{bag_classifier}
\hat{Y}=g(\sigma(f(\bm{x}_{1}),f(\bm{x}_{2}),...,f(\bm{x}_{K})))
\end{align}
% Depending on the choice of functions $f(\cdot), \sigma(\cdot), g(\cdot)$, there are two main MIL approaches: (i) instance-level approach and (ii) embedding-level approach. In the instance-level approach, the function $f(\cdot)$ is an instance-level classifier that returns a score for each instance, and the symmetric function $\sigma(\cdot)$ aggregates these instance scores to obtain a bag score. The function $g(\cdot)$ is the identity function. In the embedding-level approach, the function $f(\cdot)$ transforms each instance into a lower-dimensional embedding, and the identity function $\sigma(\cdot)$ derives a representation of the bag from the instance embeddings. The function $g(\cdot)$ classifies the bag based on this representation.
\subsection{Attention-based Deep Multiple Instance Learning}
~\cite{ITW:2018} proposed a method that parameterizes the three-stage transformation operations for classifying bags using neural networks in an embedding-level approach. First, for the k-th instance $\bm{x}_k$, a neural network $f_{\psi}(\cdot)$ with parameters $\psi$ is used to transform it into an M-dimensional embedding $\bm{h}_k = f_{\psi}(\bm{x_k}) (\in \mathbb{R}^{M})$. The K embedded instances are represented as $H = \lbrace \bm{h}_1, \bm{h}_2, ..., \bm{h}_K \rbrace$. An attention-based MIL pooling is applied to $H$. In this MIL pooling, a weighted average of instances is used, where the weights are determined by a neural network. The bag representation $\mathbf{z}$ obtained through MIL pooling is expressed as in (\ref{MIL_pooling1}) and (\ref{MIL_pooling2}).
\begin{gather}
\label{MIL_pooling1}
\mathbf{z} = \sum_{k=1}^{K}a_{k}\mathbf{h}_k\\
\label{MIL_pooling2}
a_k = \frac{\mathrm{exp}\lbrace\mathbf{w}^{\top}(\mathrm{tanh}(\mathbf{V}\mathbf{h}^{\top}_{k})\odot\mathrm{sigm}(\mathbf{U}\mathbf{h}^{\top}_{k}))\rbrace}{\sum_{j=1}^{K}\mathrm{exp}\lbrace\mathbf{w}^{\top}(\mathrm{tanh}(\mathbf{V}\mathbf{h}^{\top}_{j})\odot\mathrm{sigm}(\mathbf{U}\mathbf{h}^{\top}_{j}))\rbrace}
\end{gather}
$\mathbf{w}\in\mathbb{R}^{L\times1}$, $\mathbf{V}\in\mathbb{R}^{L\times M}$, and $\mathbf{U}\in\mathbb{R}^{L\times M}$ are learnable parameters. $\mathrm{tanh}(\cdot)$ is the hyperbolic tangent function, $\mathrm{sigm}(\cdot)$ is the sigmoid function, and $\odot$ denotes the Hadamard product. The weight $a_k$ assigned to each instance falls within $a_k \in [0, 1]$ due to the application of the softmax function. This structure, shown in \ref{MIL_pooling1} and \ref{MIL_pooling2}, calculates the similarity between instances. MIL pooling incorporates a gating mechanism proposed by ~\cite{10.5555/3305381.3305478}, addressing the issue of limited expressiveness caused by $\mathrm{tanh}(x)$ being almost linear in $x \in [-1, 1]$. Finally, the bag representation $\mathbf{z}$ is fed into a neural network $g_{\phi}(\cdot)$ with parameters $\phi$ to output the probability $\theta(X)\in[0,1]$ that $Y=1$. $\theta(X)$ is the parameter of the Bernoulli distribution that follows the bag label, and the model is trained by optimizing the log-likelihood function. In other words, binary cross-entropy is used as the loss function. Since the dataset we use has a bias towards negative labels for slice images, weighted binary cross-entropy is used to address the imbalance. Let the total number of bags be $N$, the number of positive bags be $n_p$, and the number of negative bags be $n_n$. The loss function $L$ is expressed by (\ref{loss1}) and (\ref{loss2}).
\begin{gather}
L=-(w_p\cdot Y\cdot \log \theta(X)+w_n\cdot (1-Y)\cdot \log (1-\theta(X)))\label{loss1}\\ w_p=\frac{N}{n_p},w_n=\frac{N}{n_n}\label{loss2}
\end{gather}
Since the series of transformations is differentiable, the model can be trained end-to-end using backpropagation.
\subsection{Self-supervised Learning Using Patch Images}
In this study, we divide brain CT slice images into grid-based patch images, treating each patch as an instance and the collection of patches after division as a bag. Each instance has an image size of 32×32 pixels, and there are 256 instances within each bag. The large number of instances within a bag complicates optimization in multi-instance learning. To address this, we propose a method for self-supervised learning using patch images to enhance the feature extractor $f_{\psi}(\cdot)$ for instances. This self-supervised learning combines contrastive learning, which compares features between patch images, with a reconstruction task that targets patch images transformed by rotation and flipping.

For the self-supervised learning module, we adopt the Preservational Contrastive Representation Learning (PCRL) method, as proposed by ~\cite{9710131}. Although based on contrastive learning, this approach incorporates an additional reconstruction task, enabling the encoding of more information into the representation. The reconstruction task, when learned simultaneously with contrastive learning, addresses the challenge of capturing global features—which is a common issue when applying contrastive learning to medical images—by also capturing local features closely related to the input.

The method introduces two key mechanisms: Transformation-conditioned Attention and Crossmodel Mixup. Transformation-conditioned Attention encodes the transformation vector into high-level representations using an attention mechanism. Crossmodel Mixup creates a hybrid encoder by mixing feature representations from a standard encoder and a momentum encoder, allowing for a variety of reconstruction tasks.

In our study, we modify the data augmentations used in PCRL to better suit brain hematoma CT images. Specifically, we apply multi-instance learning to brain CT slice images by dividing them into 256 patches, aiming to recognize hematoma based on texture information. To preserve as much texture information as possible, we employ crop, rotation, horizontal flipping, and cutout as data augmentations in our method (see Figure \ref{fig:DA}). Cropping involves padding the patch images with zeros by 4 pixels on each side, followed by resizing to 32×32 pixels. Cutout randomly masks three 4×4 pixel areas with zeros. Rotation is applied randomly within a range of $10^\circ$, and horizontal flipping is performed with a probability of 0.5. As loss functions, contrastive loss is used for contrastive learning, and mean squared error is used for the reconstruction task. The weights of the encoder after training are then used as the feature extractor for instances in Deep MIL.

\section{Experiments}\label{experiments}
\begin{table*}[tb]
 \centering
  \caption{Performance evaluation for hematoma detection (dataset 2-1)}
   \label{tb:hematoma_result}
  \begin{tabular}{c|lcccc}\hline
   \multicolumn{2}{c}{}& acc & prec & rec & f1  \\ \hline\hline
    &(A) Baseline& .842&.507& .846 &.463  \\ \hline
    \multirow{3}{*}{\rotatebox[]{90}{\small transfer}}&(B) Supervised Learning (ImageNet)& .864&.549& \textbf{.889}& .679\\
    &(C) \textbf{Ours 1}: Self-Supervised Learning(dataset 1)& .870& .564& .858& .681\\
    &(D) \textbf{Ours 2}: Self-Supervised Learning(dataset 2-1)&\textbf{.889} & \textbf{.607}&.883 &\textbf{.720}\\ \hline
\multirow{3}{*}{\rotatebox[]{90}{\small fine-tune}}&(B) Supervised Learning (ImageNet)& .883 &.591& \textbf{.889}&.710\\ 
&(C) \textbf{Ours 1}: Self-Supervised Learning(dataset 1)& \textbf{.886}&\textbf{.601}& .871& \textbf{.711}\\ 
&(D) \textbf{Ours 2}: Self-Supervised Learning(dataset 2-1)& .880&.585 & .883 & .704\\ \hline
\end{tabular}
\end{table*}

\begin{table*}[tb]
 \centering
  \caption{Performance evaluation of hypodensity classification (dataset 2-2) }
   \label{tb:hypodensity_result1}
  \begin{tabular}{c|lcccc}\hline
   \multicolumn{2}{c}{}& acc & prec & rec & f1  \\ \hline\hline
   &(A) Baseline& .792&.298& .853 &.441 \\ \hline
    \multirow{3}{*}{\rotatebox[]{90}{\small transfer}}&(B) Supervised Learning (ImageNet)& .865&.414& \textbf{.955}& .578\\
    &(C) \textbf{Ours 1}: Self-Supervised Learning(dataset 1)& .865&.413& .943& .575\\
    &(D) \textbf{Ours 2}: Self-Supervised Learning(dataset 2-2)&\textbf{.880} & \textbf{.440}&.921 &\textbf{.596}\\ \hline

    \multirow{3}{*}{\rotatebox[]{90}{\small fine-tune}}&(B) Supervised Learning (ImageNet)& \textbf{.896}&\textbf{.480}& .955&\textbf{.639}\\ 
    &(C) \textbf{Ours 1}: Self-Supervised Learning(dataset 1)& .850& .386&.932& .546\\ 
    &(D) \textbf{Ours 2}: Self-Supervised Learning(dataset 2-2)& .876&.436& \textbf{.966} & .601\\ \hline
    
  \end{tabular}
\end{table*}

\begin{table*}[t]
 \centering
  \caption{Performance evaluation of hypodensity classification (dataset 2-3)}
   \label{tb:hypodensity_result2}
  \begin{tabular}{c|lcccc}\hline
   \multicolumn{2}{c}{}& acc & prec & rec & f1 \\ \hline\hline
    &(A) Baseline&.788&.226& .750&.347 \\ \hline
    \multirow{3}{*}{\rotatebox[]{90}{\small transfer}}&(B) Supervised Learning (ImageNet)& .825&.288& .894& .435\\
    &(C) \textbf{Ours 1}: Self-Supervised Learning(dataset 1)& .834&.301& \textbf{.907}& .452\\
    &(D) \textbf{Ours 2}: Self-Supervised Learning(dataset 2-3)&\textbf{.853} &\textbf{.318}&.828&\textbf{.459}\\ \hline

    \multirow{3}{*}{\rotatebox[]{90}{\small fine-tune}}&(B) Supervised Learning (ImageNet)& \textbf{.853} &\textbf{.330}& \textbf{.921}&\textbf{.486}\\ 
    &(C) \textbf{Ours 1}: Self-Supervised Learning(dataset 1)& .848& .314&.855& .459\\ 
    &(D) \textbf{Ours 2}: Self-Supervised Learning(dataset 2-3)& .842&.313 & \textbf{.921} & .468\\ \hline
    
  \end{tabular}
\end{table*}

% \begin{table*}[t]
%  \centering
%   \begin{tabular}{c|cc|cc}\hline\hline
%     &\multicolumn{2}{c|}{Hematoma detection}&\multicolumn{2}{c}{Marker classification} \\ \hline
%     Methods & accuracy & f1 & accuracy & f1 \\ \hline
%     Baseline &.842&.463 &.788 &.347 \\
%     Supervised learning& .864&.679&.825 &.435\\
%     Self-supervised learning (\textbf{Ours})& \textbf{.889}&\textbf{.720} & \textbf{.853} &\textbf{.459}\\ \hline
%   \end{tabular}
% \end{table*}

\begin{figure*}[t]
\begin{center}
\includegraphics[width=12cm]{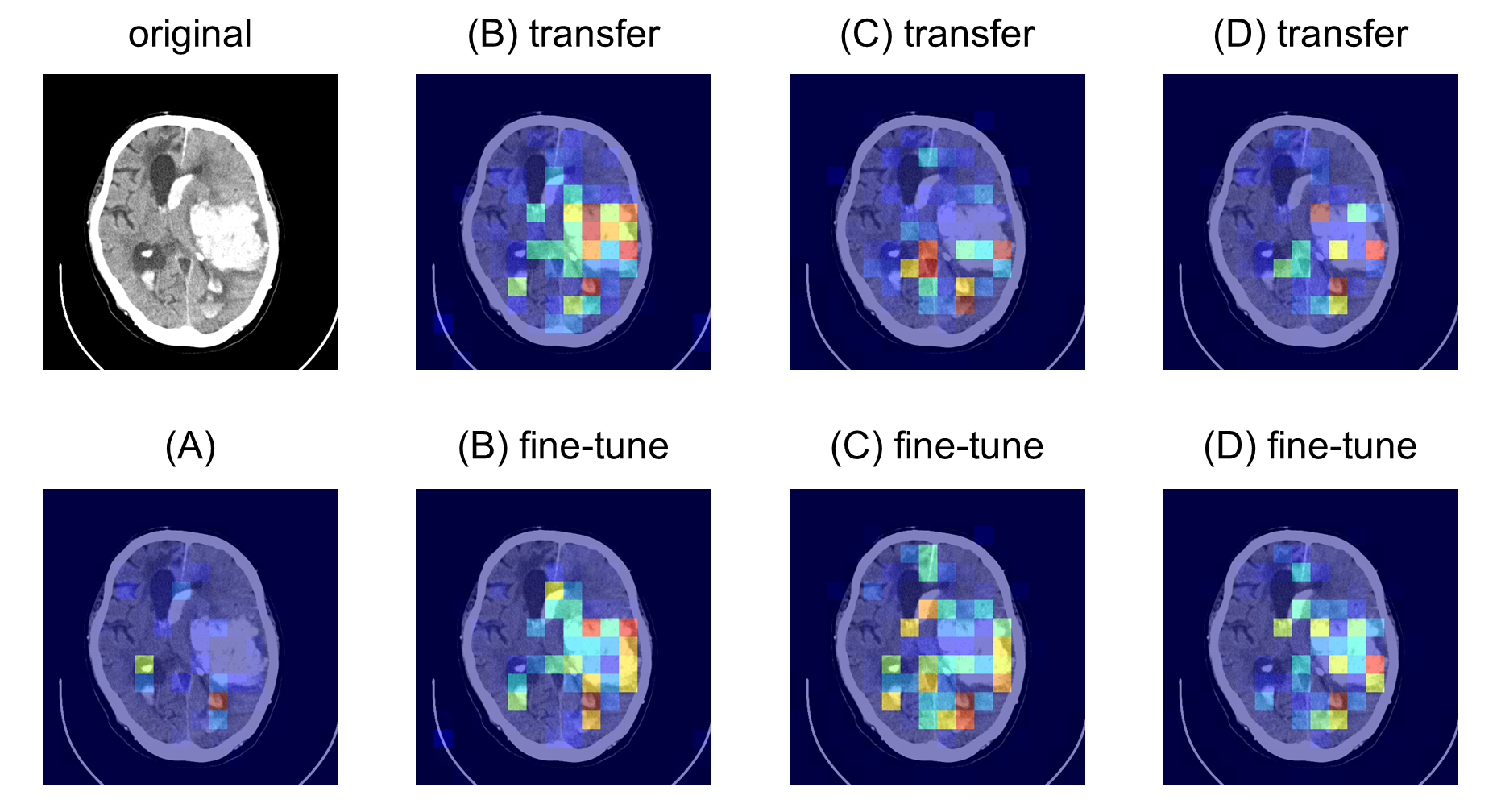}
\caption{attention maps for hematoma detection}
\label{fig:attention_map_hematoma}
\end{center}
\end{figure*}

\begin{figure*}[t]
\begin{center}
\includegraphics[width=12cm]{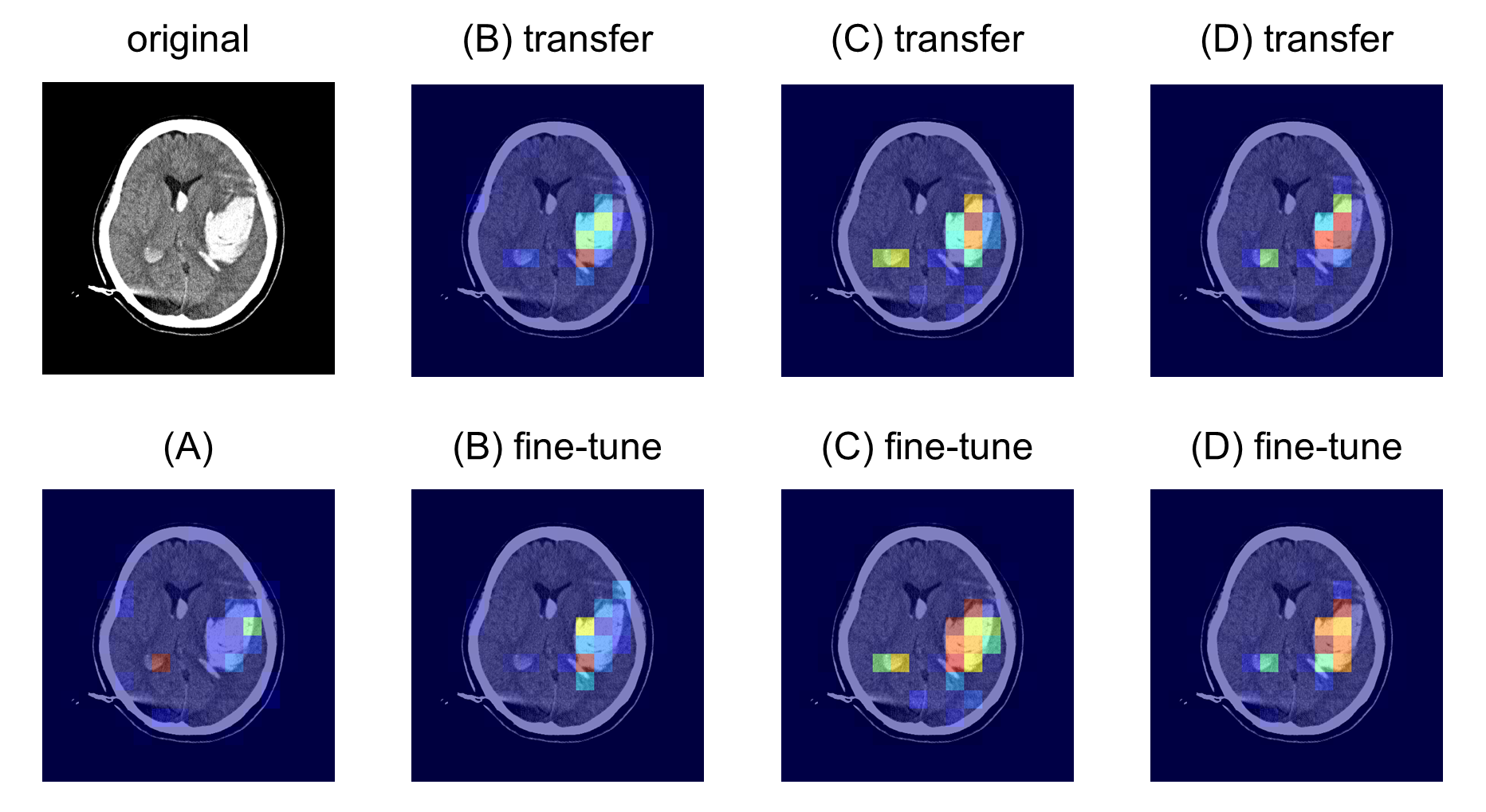}
\caption{attention map for hypodensity
(dataset 2-2) classification}
\label{fig:attention_map_hypodensity1}
\end{center}
\end{figure*}

\begin{figure*}[t]
\begin{center}
\includegraphics[width=12cm]{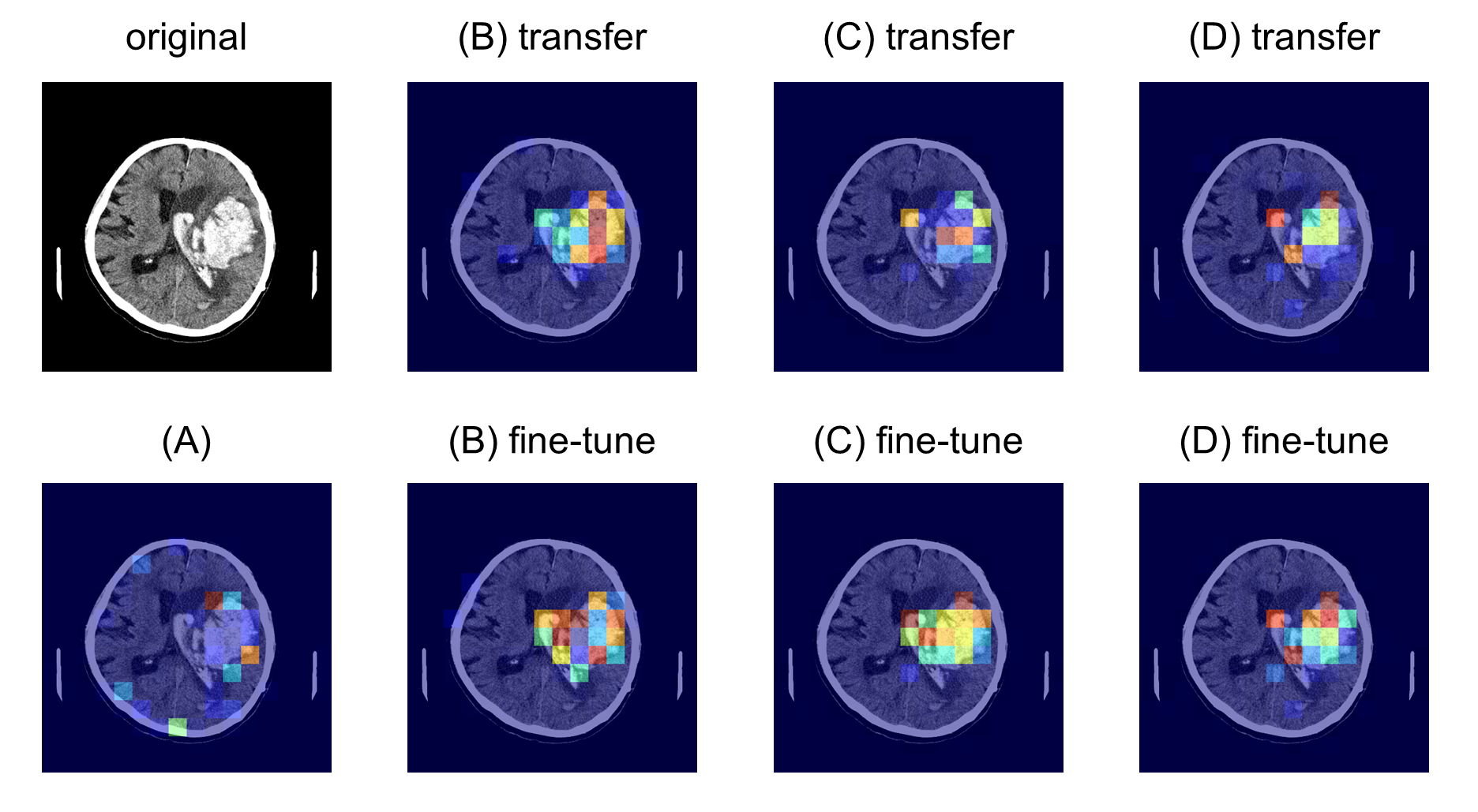}
\caption{attention map for hypodensity (dataset 2-3) classification}
\label{fig:attention_map_hypodensity2}
\end{center}
\end{figure*}
\subsection{CT Image Data}
CT images are stored in DICOM format with a size of 512×512. The pixel values are in Hounsfield units (HU). Before saving as PNG format, windowing is applied as a preprocessing step. The window range determined by specialists is used to select the target HU range and adjust the contrast. If the window width is described in the DICOM file metadata, that value is used. Let $a$ be the lower window limit (unit: HU), $b$ be the upper window limit (unit: HU), and $I_{HU}(i,j)$ be the pixel value at position $(i,j)$ of the slice image before contrast adjustment. The pixel value $I(i,j)$ after contrast adjustment is given by  (\ref{HU_contrast}).
\begin{align}
\label{HU_contrast}
I(i,j)
 =
  \begin{cases}
    0& \mathrm{if}\;I_{HU}(i,j)<a \\
    \frac{I_{HU}(i,j)-a}{b-a}\times 255&  \mathrm{if}\;a\leq I_{HU}(i,j)\leq b\\
    255& \mathrm{if}\;I_{HU}(i,j)>b
  \end{cases}
\end{align}
\subsection{Datasets}
In this study, we used two datasets: the RSNA dataset\footnote{https://www.kaggle.com/c/rsna-intracranial-hemorrhage-detection\label{rsna_url}} (Dataset 1) and in-house CT scan data (Dataset 2)\\
\indent Dataset 1 is a publicly available dataset, which was used solely for the pre-training of the instance feature extractor in this study. Each slice is labeled by specialists for five possible overlapping subtypes of hematomas: intraparenchymal, intraventricular, subarachnoid, subdural, and epidural. However, Dataset 2 does not include hematomas classified as subarachnoid, subdural, or epidural. Therefore, slices classified under any of these subtypes were excluded. The dataset consists of over 750,000 slices, and using all slices for training would require a significant amount of time. Hence, a randomly selected subset was used for training.\\
\indent Dataset 2 is a private dataset, which was used in this study for both the pre-training of the instance feature extractor and for the detection of hematomas and brain hematoma markers (hypodensity) in multiple instance learning. Each slice is annotated by specialists for four possible overlapping brain hematoma markers (hypodensity, margin irregularity sign, blend sign, and fluid level). For the hematoma detection task, if at least one of the four brain hematoma markers is positive, the slice is labeled as having a hematoma. If none of the markers are positive, the slice is labeled as not having a hematoma (Dataset 2-1). For the hypodensity detection task, datasets were created by dividing slices based on the presence or absence of hypodensity. In this case, two versions of the dataset were created: one excluding data where hematomas are present but hypodensity is negative (Dataset 2-2), and the other including such data (Dataset 2-3).\\
\indent The breakdown of each dataset (slices) is shown in Table \ref{tb:dataset_detail}. To apply multiple instance learning, each 512×512 slice image was divided into 32×32 patch images, resulting in 256 patch images per slice. When patch images are input into the network, the pixel values are normalized to the range of 0 to 1. Although the patch images are 32×32×1 grayscale images, the network processes RGB images as input. Thus, the pixel values of the grayscale images were pseudo-converted into 32×32×3 RGB images by replicating the grayscale values across all three channels of the RGB images.
\begin{table}[H]
 \begin{center}
   \caption{Breakdown of datasets (Slices)}
    \label{tb:dataset_detail}
  \begin{tabular}{ccccc}\hline\hline
    & Data Group & Negative & Positive & Total\\ \hline
    \multirow{2}{*}{Dataset 1} & train & 3786 & 214 & 4000\\
    & valid & 953 & 47 & 1000\\ \hline
    \multirow{3}{*}{Dataset 2-1} & train & 6709 & 1363 & 8072\\
    & valid & 839 & 170 & 1009\\ 
    & test & 847 & 163 & 1010\\ \hline
    \multirow{3}{*}{Dataset 2-2} & train & 6712 & 677 & 7389\\
    & valid & 847 & 76 & 923\\ 
    & test & 836 & 89 & 925\\ \hline
    \multirow{3}{*}{Dataset 2-3} & train & 7394 & 678 & 8072\\
    & valid & 921 & 88 & 1009\\ 
    & test & 934 & 76 & 1010\\ \hline
  \end{tabular}
 \end{center}
\end{table}
\subsection{Experimental Procedure}
\begin{figure}[H]
\begin{center}
\includegraphics[width=8cm]{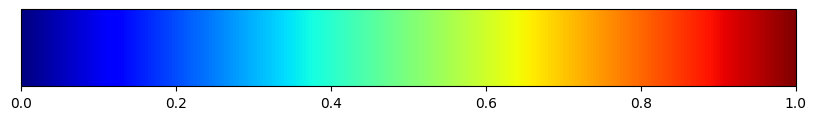}
\caption{Jet Colormap}
\label{fig:color_map_jet}
\end{center}
\end{figure}

To verify the effectiveness of our method, we conducted comparative experiments under the following conditions:
\begin{enumerate}
\renewcommand{\labelenumi}{(\Alph{enumi})}
 \item Using only attention-based deep multiple instance learning. This serves as the baseline method for our study. LeNet5 is used as the Encoder, and the weights are initialized randomly.
 \item Using encoder pre-trained through supervised learning on ImageNet. VGG11 is used as the Encoder.
 \item VGG11 pre-trained in ImageNet is further trained in self-supervised learning using dataset 1. (\textbf{Ours 1})
 \item VGG11 pre-trained in ImageNet is further trained in self-supervised learning using the same dataset used for multiple instance learning (excluding test data). (\textbf{Ours 2})
\end{enumerate}
\indent For conditions (B) to (D), we compare cases where the VGG weights are frozen (transfer learning) and not frozen (fine-tuning) during the downstream multiple instance learning stage. Additionally, under conditions (A) to (D), we perform hematoma detection and classification of brain hematoma markers (hypodensity). For hematoma detection, Dataset 2-1 is used, and for hypodensity classification, we use both Dataset 2-2 and Dataset 2-3 for multiple instance learning. The model from the epoch with the minimum validation loss during training is used as the trained model. Test data is used for performance evaluation of hematoma detection and hypodensity classification. The performance metrics include accuracy, precision, recall, and F1-score. Regarding the attention weights $a_k\in [0,1]$ assigned to each instance, the jet colormap (Figure \ref{fig:color_map_jet}) is used to visualize the attention map. This visualization allows easy identification of the instances that the model focuses on. The parameters used during pre-training with self-supervised learning and those used during downstream multiple instance learning are shown in Table \ref{tb:parameter}. All experiments were conducted using an NVIDIA GeForce RTX 4090 GPU, Python 3.11.3, and Pytorch 2.0.1+cu118.
\begin{table}[H]
 \centering
  \caption{Parameters}
   \label{tb:parameter}
  \begin{tabular}{ccc}\hline\hline
   & Item & Value \\ \hline
   \multirow{6}{*}{Pre-training with SSL}&encoder & vgg11 \\
    &batch size & 256 \\ 
    &learning rate & 0.005 \\
   &optimizer & sgd \\ 
   &weight decay & 0.0001\\
    &momentum & 0.9 \\ \hline
    \multirow{4}{*}{Deep MIL}&learning rate & 0.000001 \\
   &optimizer & Adam \\ 
   &weight decay & 0.00001\\
    &$\beta_{1},\beta_{2}$ & 0.9, 0.999 \\ \hline
  \end{tabular}
\end{table}

\subsection{Results}
The evaluation values and attention maps for hematoma detection under each condition are shown in Table \ref{tb:hematoma_result} and Figure \ref{fig:attention_map_hematoma}. The evaluation values and attention maps for hypodensity (dataset 2-2) classification under each condition are shown in Table \ref{tb:hypodensity_result1} and Figure \ref{fig:attention_map_hypodensity1}. The evaluation values and attention maps for hypodensity (dataset 2-3) classification under each condition are shown in Table \ref{tb:hypodensity_result2} and Figure \ref{fig:attention_map_hypodensity2}. The attention map uses slices where the correct label is positive and the slice was correctly predicted as positive under each condition.

\subsection{Discussion}
First, we examine the hematoma detection performance(Table \ref{tb:hematoma_result}). For the proposed methods (C) and (D), both transfer and fine-tune outperform the baseline method (A) across all evaluation metrics. However, to verify the effect of self-supervised learning using brain hematoma patch images, it is also crucial to compare with (B), which uses VGG pre-trained on ImageNet. In the case of transfer learning, the proposed method (D) surpasses (B) in all evaluation metrics except recall. Notably, focusing on the F-score, considering the imbalanced data, an improvement of 4.1\% over (B) is observed. Even for the recall, where performance decreased, the difference is minimal. On the other hand, for fine-tuning, (C) shows improvement in accuracy, precision, and F1, though the increases are slight. We also examine the attention map(Figure \ref{fig:attention_map_hematoma}). For the baseline method (A), it is observed that the large hematoma on the right side is barely captured. In contrast, (B) has higher Attention values covering the entire hematoma, indicating more accurate detection compared to (A). For (C) and (D), compared to (A), parts of the hematoma are captured, but not the whole.

Next, we examine the classification performance for hypodensity (Dataset 2-2)(Table \ref{tb:hypodensity_result1}). Both transfer learning and fine-tuning show that the proposed methods (C) and (D) surpass the baseline method (A) across all evaluation metrics. Furthermore, compared to the hematoma detection results, the performance improvement of the proposed methods over the baseline (A) is more significant. Similar to the hematoma detection, we compare with (B). In the case of transfer learning, the proposed method (D) shows a 1.8\% improvement in the F-score over (B). For fine-tuning, the proposed method (D) shows a 1.1\% improvement in recall, but (B) outperforms in other evaluation metrics. Examining the attention map(Figure \ref{fig:attention_map_hypodensity1}), the baseline method (A) focuses only on part of the boundary of the hematoma on the right side of the slice, failing to capture the entire hematoma. Compared to (A), (B) through (D) show better focus on the hematoma.

Additionally, we examine the classification performance for hypodensity (Dataset 2-3)(Table \ref{tb:hypodensity_result2}). Both transfer and fine-tune show that the proposed methods (C) and (D) surpass the baseline method (A) across all evaluation metrics. Comparing with (B), in the case of transfer learning, the proposed method (D) shows a 2.4\% improvement in the F-score over (B). For fine-tuning, the proposed method (D) matches (B) in recall, but (B) outperforms in other evaluation metrics. Examining the attention map(Figure \ref{fig:attention_map_hypodensity2}), the baseline method (A) focuses only on part of the boundary of the hematoma on the right side of the slice and mistakenly focuses on areas near the skull unrelated to the hematoma. Compared to (A), (B) through (D) show better focus on the hematoma.

Common to the results of hematoma detection and hypodensity classification, the precision values are significantly lower than the recall values, resulting in lower F-scores. This could be due to the use of a weighted loss function that prioritizes recall. In medical image classification, recall is crucial; hence, while maintaining high recall, improving precision is a key challenge. Comparing transfer learning and fine-tuning, although transfer learning shows performance improvement over (B), fine-tuning did not yield the expected performance. In fine-tuning, retraining the instance feature extractor in the multiple instance learning process might have been adversely affected by the large number of instances (256). In transfer learning, the weights of the pre-trained model are fixed, leading to performance improvement, suggesting that self-supervised learning of patch images was effective. Comparing (C), which used Dataset 1 (RSNA dataset) for self-supervised learning, and (D), which used the same dataset as multiple instance learning, transfer learning shows a slight decrease in performance in the former case. However, focusing on the classification results for hypodensity (Dataset 2-3), (C) shows a 1.7\% improvement in the F-score over (B). This suggests the potential of learning useful features even when using a different dataset for self-supervised learning than that used for multiple instance learning. The RSNA dataset used only a small portion of the data due to training time constraints, indicating the possibility of further performance improvement by expanding the dataset.

\section{Conclusions}\label{Conclusion}
In this study, we propose using a pre-trained model with self-supervised learning for the multi-instance learner as a downstream task. Specifically, we apply self-supervised learning on patch images as the instance feature extractor in Deep MIL for the classification of brain hematoma images. The results showed that our method improved performance compared to the baseline Deep MIL without self-supervised learning, indicating that providing information through self-supervised learning on patch images is beneficial. Additionally, in transfer learning, our approach demonstrated better classification performance even compared to supervised learning using ImageNet, suggesting that self-supervised learning effectively captured useful features in patch images. Furthermore, attention visualization revealed that the proposed method successfully captured hematomas that were not detected by the baseline. Looking forward, three main directions are suggested: first, to conduct evaluation experiments by expanding the dataset used for self-supervised learning on patch images; second, to explore approaches for improving precision; and third, to evaluate the performance of the proposed method on tasks other than the classification of brain hematoma images.
\clearpage
\bibliography{aaai25}

\end{document}